\documentclass{article}

\usepackage{microtype}
\usepackage{graphicx}
\usepackage{subfigure}
\usepackage{booktabs} 
\usepackage{subcaption}
\usepackage{hyperref}
\usepackage{multirow}

\usepackage[accepted]{icml2025}

\usepackage{amsmath}
\usepackage{amssymb}
\usepackage{mathtools}
\usepackage{amsthm}

\usepackage{graphicx}
\usepackage{dirtree}
\usepackage{algorithm}
\usepackage{hyperref}
\makeatletter
\renewcommand{\ALG@name}{List}
\makeatother

\usepackage[T1]{fontenc}
\usepackage[utf8]{inputenc}
\usepackage[USenglish]{babel}
\usepackage{algorithmic}
\usepackage{float}
\usepackage{booktabs}

\PassOptionsToPackage{numbers,sort&compress}{natbib}
\usepackage{natbib}
\usepackage[numbers]{natbib}

\usepackage[utf8]{inputenc} 
\usepackage[T1]{fontenc}    
\usepackage{hyperref}       
\usepackage{url}            
\usepackage{booktabs}       
\usepackage{amsfonts}       
\usepackage{nicefrac}       
\usepackage{microtype}      
\usepackage{xcolor}         
\usepackage{booktabs}
\usepackage{tabularx} 
\newcommand{\bh}{\mathbf{h}}

\usepackage[T1]{fontenc}
\usepackage[utf8]{inputenc}
\usepackage[USenglish]{babel}
\usepackage{algorithmic}
\usepackage[capitalize,noabbrev]{cleveref}

\theoremstyle{plain}

\theoremstyle{definition}

\theoremstyle{remark}

\usepackage[textsize=tiny]{todonotes}

\title{BioLangFusion:  Multimodal Fusion of DNA, mRNA, and Protein Language Models}

\begin{document}

\twocolumn[
\icmltitle{BioLangFusion:  Multimodal Fusion of DNA, mRNA, and Protein Language Models}




\icmlsetsymbol{equal}{*}

\begin{icmlauthorlist}
\icmlauthor{Amina Mollaysa}{yyy}
\icmlauthor{Artem Moskalev}{yyy}
\icmlauthor{Pushpak Pati}{yyy}
\icmlauthor{Tommaso Mansi}{yyy}
\icmlauthor{Mangal Prakash}{equal,yyy}
\icmlauthor{Rui Liao}{equal,yyy}
\end{icmlauthorlist}

\icmlaffiliation{yyy}{Johnson $\&$ Johnson Innovative Medicine}

\icmlcorrespondingauthor{Amina Mollaysa}{maminanm@its.jnj.com}

\icmlkeywords{Machine Learning, ICML}

\vskip 0.3in
]
\printAffiliationsAndNotice{\icmlEqualContribution}



\begin{abstract}

We present BioLangFusion, a simple approach for integrating pre-trained DNA, mRNA, and protein language models into unified molecular representations. Motivated by the central dogma of molecular biology (information flow from gene to transcript to protein), we align per-modality embeddings at the biologically meaningful codon level (three nucleotides encoding one amino acid) to ensure direct cross-modal correspondence. BioLangFusion studies three standard fusion techniques: (i) codon-level embedding concatenation, (ii) entropy-regularized attention pooling inspired by multiple-instance learning, and (iii) cross-modal multi-head attention—each technique providing a different inductive bias for combining modality-specific signals. These methods require no additional pre-training or modification of the base models, allowing straightforward integration with existing sequence based foundation models. Across five molecular property prediction tasks, BioLangFusion outperforms strong unimodal baselines, showing that even simple fusion of pre-trained models can capture complementary multi-omic information with minimal overhead.

\end{abstract}
 \section{Introduction}
 The central dogma of molecular biology—DNA is transcribed into mRNA, which is translated into protein—captures a coordinated flow of genetic information. While each stage carries unique regulatory signals, all contribute jointly to phenotype. For instance, disrupting a transcription factor binding site (DNA) can destabilizing an mRNA hairpin, and alter codon usage or amino acid identity (protein). Understanding such effects requires reasoning across this entire molecular cascade.

Foundation models (FMs) trained on single modalities have recently shown impressive capabilities in their respective domains. Existing DNA language models~\citep{ji2021dnabert, nucleotidetransformer2022, nguyen2024evo} capture regulatory sequence patterns; mRNA-focused models~\citep{zhang2023codonbert, yazdani2024helm} encode structural and post-transcriptional features; and protein models~\citep{esm2_2022, prottrans2021} excel in predicting structure and function from amino acid sequences. However, these FMs are trained on single modality, ignoring the intrinsic linkage among modalities implied by the central dogma. Emerging evidence shows that unimodal models can exhibit surprising cross-modal capabilities—for example, Evo-2~\citep{brixi2025genome}, a 40B-parameter DNA-only model, performs well on RNA and protein tasks~\citep{nguyen2024evo}, but these effects arise from massive scale and compute—resources that are simply not affordable or accessible to most researchers. Separately, DNA or protein FMs have also shown utility for mRNA-specific tasks~\citep{prakash2024bridging}, suggesting that some molecular signals do transfer even without explicit multimodal training. Still, such generalization is limited and incidental: the models do not explicitly model the flow of information across modalities. As a more pragmatic alternative, recent work shows that even simply concatenating embeddings from DNA and protein models can improve performance on downstream tasks~\citep{boshar2024are}, implying that the two modalities encode complementary—and sometimes orthogonal—biological cues.

These observations suggest that meaningful fusion of modality-specific embeddings—combining representations from DNA, RNA, and protein models—could better capture the full DNA→RNA→protein cascade. One straightforward method is direct concatenation of embeddings, but this often fails to scale effectively due to lack of alignment and limited ability to capture cross-modal interactions. Another approach, weight merging, requires foundation models to share the same architecture, tokenization, and embedding dimensions—constraints that are rarely met. Finally, fusion methods like FuseLM from natural language processing rely on knowledge distillation and retraining, which introduce substantial computational complexity.

To address these challenges, we introduce BioLangFusion, a suite of simple and modular fusion techniques that integrates pretrained DNA, RNA, and protein FMs embeddings without requiring additional training or architectural modifications. We first align embeddings at the codon level to establish biologically meaningful correspondence across modalities. BioLangFusion then studies three fusion strategies: (i) codon-level concatenation, (ii) entropy-regularized attention pooling inspired by multiple-instance learning, and (iii) cross-modal multi-head attention capturing token-level dependencies. Evaluated on five diverse molecular property prediction tasks, BioLangFusion techniques consistently improve over unimodal baselines, offering a practical method for integrating multimodal omics data spanning central dogma with modest computational overhead.

\section{Methods}
\label{seq:method}
\vspace{-0.3em}

Assume we have training set \( D = \{(\mathbf{x}_i, y_i)\}_{i=1}^N \), where \( \mathbf{x}_i = [x_{i1}, \dots, x_{iT}] \) is a an mRNA sequence with length $T$ where $x_i$ denotes nucleotide,  and \( y_i \) is the molecular property of interest. Each mRNA maps to its corresponding DNA and protein sequences via the central dogma given the start and stop codons signaling the start and end of translation. Our objective is to predict \(y_i\) by fusing the embeddings 
\(\;E_{\mathrm{DNA}},\,E_{\mathrm{RNA}},\,E_{\mathrm{Prot}}\)\ of sequence $\mathbf{x}_i$ extracted from pretrained DNA, RNA, and protein language models.
\vspace{-0.5em}
\subsection{Modality alignment}\label{sec:modality_align} 
Language models tokenize their inputs at varying biological granularities, resulting in embeddings often differ in precision. Since different pre-trained FM perform variably across tasks \citep{prakash2024bridging, boshar2024are} and there is no universal rule for model selection, we adopt well-established models representative of each modality: the \emph{Nucleotide Transformer}~\cite{dalla2023nucleotide} for DNA (6-mer tokenization), \emph{RNA-FM}~\cite{chen2022rnafm} for RNA (single nucleotide tokenization), and \emph{ESM-2}~\cite{esm2_2022} for protein (amino acid tokenization, i.e., 3-mers in nucleotide). 
\vspace{-0.5em}
\begin{equation}
\resizebox{\columnwidth}{!}{$
    E_{\mathrm{DNA}} \in \mathbb{R}^{\frac{T}{6} \times d_{\mathrm{DNA}}}; \quad
    E_{\mathrm{RNA}} \in \mathbb{R}^{T \times d_{\mathrm{RNA}}}; \quad
    E_{\mathrm{Prot}} \in \mathbb{R}^{\frac{T}{3} \times d_{\mathrm{Prot}}}
$} \nonumber
\end{equation}
where \( d_{\mathrm{DNA}}, d_{\mathrm{RNA}}, d_{\mathrm{Prot}} \) denote the embedding dimensions of the respective language models.

Consequently, embeddings derived from these models inherently vary in length and must be aligned for effective fusion, ensuring that corresponding tokens across modalities represent the same biological region and carry semantically aligned information. We use the protein frame (\( T' = \tfrac{T}{3} \) length) as the reference and map the DNA and mRNA embeddings onto shared codon-level resolution. This choice is biologically motivated: proteins are the final functional products of the central dogma, and their sequences are directly derived from coding regions in DNA and mRNA via translation. Aligning at the codon level—where each codon consists of three nucleotides encoding a single amino acid—ensures that each token corresponds to a biologically meaningful unit. To achieve this alignment, we apply transposed convolution to upsample the DNA embeddings (originally defined over 6-mers) to the codon level, preserving local context, and apply non-overlapping mean pooling to downsample mRNA embeddings to the same token resolution.
\vspace{-0.5em}
\begin{equation}
\resizebox{0.7\columnwidth}{!}{$
\begin{split}
  \tilde E_{\mathrm{DNA}}
  &= \operatorname{TConv}_{k=2,s=2}\bigl(E_{\mathrm{DNA}}\bigr)
     \;\in\;\mathbb{R}^{T'\times d_{\mathrm{DNA}}},\\
  \tilde E_{\mathrm{RNA}}
  &= \operatorname{AvgPool}_{k=3,s=3}\bigl(E_{\mathrm{RNA}}\bigr)
     \;\in\;\mathbb{R}^{T'\times d_{\mathrm{RNA}}}.
\end{split}
$}
\end{equation}
After alignment, each modality is represented by a embeddings of length \( T' \), where each location \( t \) corresponds to the same biological codon across DNA, RNA, and protein representations.
\begin{figure*}
    \centering
    \includegraphics[width=0.65\textwidth, height=3.8cm]{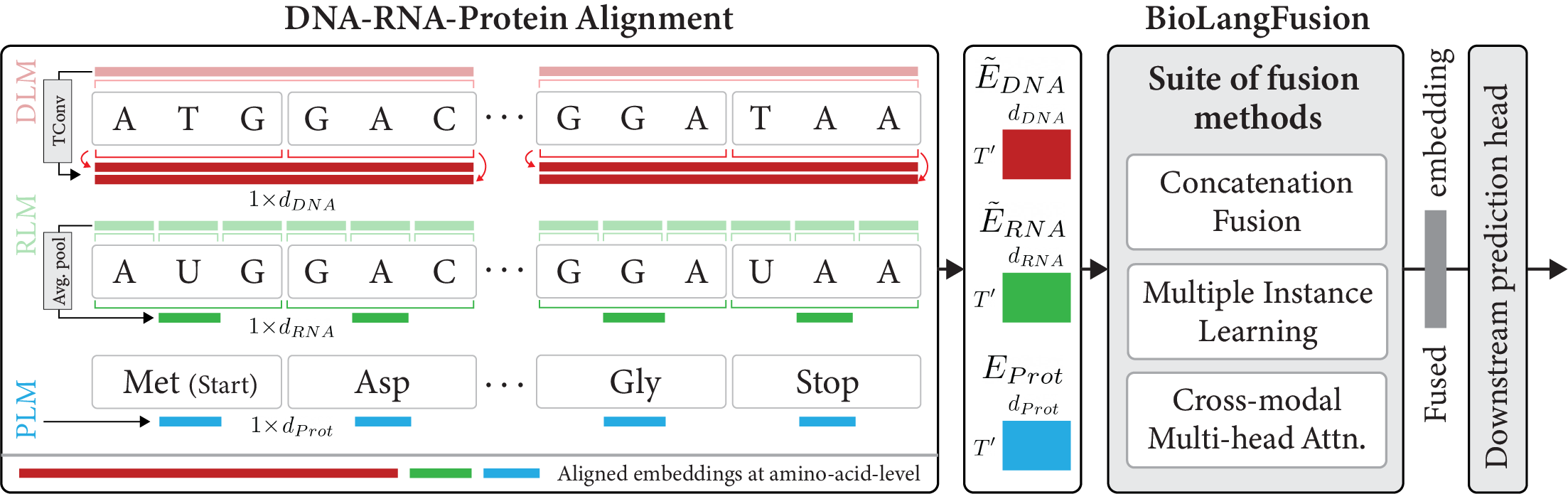}
    \caption{BioLangFusion Architecture Overview: pretrained DNA, RNA, and protein embeddings are aligned at the codon level and fused using biologically motivated strategies then passed to a prediction head for downstream molecular property prediction.}
    \label{fig:model_overview}
\end{figure*}
\vspace{-1em}
\subsection{Fusion methods}
\subsubsection{Concatenation fusion}\label{sec:concat}
Given now we have aligned embeddings:
$\tilde E_{\mathrm{DNA}}$, $\tilde E_{\mathrm{RNA}}$, $E_{\mathrm{Prot}}$, 
a straightforward fusion strategy is to \emph{concatenate} the feature vectors from each modality at every aligned position along feature dimension.
However, due to differences in embedding dimensionality across modalities, such concatenation can lead to imbalanced representations, where the modality with the largest embedding dimension may dominate the fused vector (see Table~\ref{tab:foundation_models}).
To address this, we apply a modality-specific learnable MLP to project the DNA embedding to a lower-dimensional space of size \(d'_{\mathrm{DNA}}\) prior to concatenation since $d_{DNA}$ is significantly larger than $d_{RNA}$ and $d_{Prot}$.
\vspace{-0.5em}
\begin{equation}
\resizebox{\columnwidth}{!}{$
Z_{\mathrm{concat}}(t)
= \bigl[\; \mathrm{MLP}(\tilde E_{\mathrm{DNA}}[t]) \,\Vert\, \tilde E_{\mathrm{RNA}}[t] \,\Vert\, E_{\mathrm{Prot}}[t] \;\bigr],
\quad t = 1, \dots, T'\nonumber
$}
\end{equation}
\vspace{-0.5em}
resulting in \( Z_{\mathrm{concat}} \in \mathbb{R}^{T' \times (d'_{\mathrm{DNA}} + d_{\mathrm{RNA}} + d_{\mathrm{Prot}})} \).

This method preserves the full granularity of all modalities without imposing constraints on their interactions, allowing the prediction head to learn informative feature combinations. However, as simple and intuitive as this approach is, it treats all modalities equally at every position. As a result, the model cannot adaptively down-weight noisy or less informative modality unless this behavior is implicitly learned by the prediction head. Moreover, the dimensionality of \( Z_{\mathrm{concat}} \) increases linearly with the number of modalities, which leads to higher computational and memory costs. These limitations motivate our attention-based fusion alternative, which learns to softly weight each modality based on its relevance to the task at hand.

\subsubsection{Multiple Instance Learning (MIL) with Entropy Regularizer}\label{sec:mil}
Inspired by multiple instance learning in computer vision \cite{ilse2018attention}, we treat the three modality-specific embeddings as a bag of instances, where each instance (modality) may contribute differently to the final prediction. We introduce a lightweight gated attention mechanism  that \emph{dynamically} weights each modality based on its relevance to the task.

For simplicity, we denote the aligned embeddings as:
$\tilde{ E}_{m}\in\mathbb R^{T'\times d_{m}},\quad m\in\{\mathrm{DNA},\mathrm{RNA},\mathrm{Prot}\}$
 where each modality has its own feature dimension. To apply attention pooling across modalities, we first project each embedding to a shared latent space  $H_{m}\in \mathbb R^{T'\times d}$, then apply gated attention:
\begin{equation}
Z_{\textit{fused}} = \sum_{m \in \{\mathrm{DNA}, \mathrm{RNA}, \mathrm{Prot}\}} \alpha_m H_m,
\end{equation}
where the attention weights \( \alpha_m \) are computed from a mean-pooled summary $ \bar{\bh}_m= \frac{1}{T'}\sum_{t=1}^{T'}H_m[t]$ representation of each modality. We adopt this formulation of sequence-level attention in place of token-level attention, which was empirically less effective in our setting (see Table~\ref{Tab:ablation}).
\begin{equation}\label{eq:gated_attention}
\resizebox{\columnwidth}{!}{$
  \alpha_m 
  = \frac{
    \exp\bigl(W^\top[\tanh(V_m\bar{\bh}_m + \mathbf b_m)
      \odot \sigma(U_m\bar{\bh}_m + \mathbf c_m)]\bigr)
  }{
    \sum_{i\in\{\mathrm{DNA},\mathrm{RNA},\mathrm{Prot}\}}
    \exp\bigl(W^\top[\tanh(V_i\bar{\bh}_i + \mathbf b_i)
      \odot \sigma(U_i\bar{\bh}_i + \mathbf c_i)]\bigr)
  }\nonumber$
  }
\end{equation}
 \( U_m, V_m \in \mathbb{R}^{d_{\mathrm{attn}} \times d} \), \( \mathbf{b}_m, \mathbf{c}_m \in \mathbb{R}^{d_{\mathrm{attn}}} \), and \( W \in \mathbb{R}^{d_{\mathrm{attn}}} \) are learnable parameters, \( \odot \) denotes element-wise multiplication and \( \sigma(\cdot) \) is the sigmoid function. This gated attention combines the bounded non-linearity of \( \tanh \) with the smooth gating of \( \sigma \), enhancing flexibility and mitigating \( \tanh \) saturation~\cite{ilse2018attention}.

The resulting fused embedding matrix \( Z_{\textit{fused}} \) is then passed to the downstream prediction head. This ``one-attention-per-sequence'' design is lightweight yet enables the model to \emph{dynamically} determine, for each downstream task, how much information to retain from DNA, mRNA, and protein modalities.
\vspace{-0.3em}
\paragraph{Entropy regularization.}
During training, we observe in some case, the model often struggles to learn diverse attention weights. To encourage the model to move away from such nearly uniform solutions, we add a negative entropy term to the loss function. The attention entropy over a mini-batch \( \mathcal{D}_i \) is defined as:
\begin{equation}
\resizebox{\columnwidth}{!}{$
   H_{\text{attn}}(\boldsymbol{\alpha})
   = -\frac{1}{|\mathcal{D}_i|} \sum_{(\mathbf{x}, \mathbf{y}) \in \mathcal{D}_i}
   \sum_{m \in \{\mathrm{DNA}, \mathrm{RNA}, \mathrm{Prot}\}} \alpha_m \log \alpha_m
   $}\nonumber
\end{equation}
where \( \boldsymbol{\alpha} = [\alpha_{\mathrm{DNA}}, \alpha_{\mathrm{RNA}}, \alpha_{\mathrm{Prot}}] \) are the modality-level attention weights. This regularization, $\lambda  H_{\text{attn}}(\boldsymbol{\alpha})$ is added to the main loss function where $\lambda$ is a tunable hyperparameter controlling the strength of the regularization. 



\subsubsection{Cross-modal Multi-head Attention}\label{sec:cross_attention}
While concatenation and attention-based pooling effectively combine modality-specific embeddings, they treat each modality independently at every aligned position. To capture more nuanced interactions between DNA, mRNA, and protein representations, we explore a cross-modal multi-head attention mechanism inspired by transformer architectures.

In contrast to previous fusion approaches, cross-modal attention allows each modality to query contextual information from all modalities jointly. This enables the model to dynamically discover and leverage cross-modality dependencies that are informative for downstream biological prediction tasks. At each position, the model can attend across the full set of aligned DNA, mRNA, and protein embeddings, identifying which signals are most relevant for the task. Specifically, after projecting all modality embeddings to a shared feature dimension \( d \), we concatenate them along the sequence length axis to form a global context representation:
\begin{equation}
C = [H_{\mathrm{DNA}}; H_{\mathrm{RNA}}; H_{\mathrm{Prot}}] \in \mathbb{R}^{3T' \times d}
\end{equation}
This context \( C \) provides the keys and values for attention, while the embedding of each modality acts as the query. Formally, for each modality, we compute:
\begin{equation}
\resizebox{0.8\columnwidth}{!}{$
Z_m = g\left(\mathrm{MultiHead}(H_m W^Q_m, C W^K_m, C W^V_m)\right)
$}
\end{equation}
where \( \mathrm{MultiHead} \) denotes the standard multi-head attention operator, and \( g(\cdot) \) includes residual and projection layers.

The updated embeddings \( Z_{\mathrm{DNA}}, Z_{\mathrm{RNA}}, Z_{\mathrm{Prot}} \) are then concatenated across the feature dimension and projected to form a unified representation \( Z \). To stabilize training and preserve modality-specific signals, we add a residual connection with modality averaging followed by layer normalization:
\begin{equation}
\resizebox{0.9\columnwidth}{!}{$
Z_{\textit{fused}} = \mathrm{LayerNorm}\left(\frac{Z_{\mathrm{DNA}} + Z_{\mathrm{RNA}} + Z_{\mathrm{Prot}}}{3} + Z\right) 
$}
\end{equation}
This flexible mechanism allows information to flow between modalities and adaptively reweighs their contributions across positions, capturing both shared and complementary biological features. By learning these interactions, the model builds a richer, more expressive representation than those derived from independent fusion strategies.
\vspace{-0.3em}

\section{Experiment}
\begin{figure}[ht]
    \centering
        \includegraphics[width=\columnwidth, height=3cm]{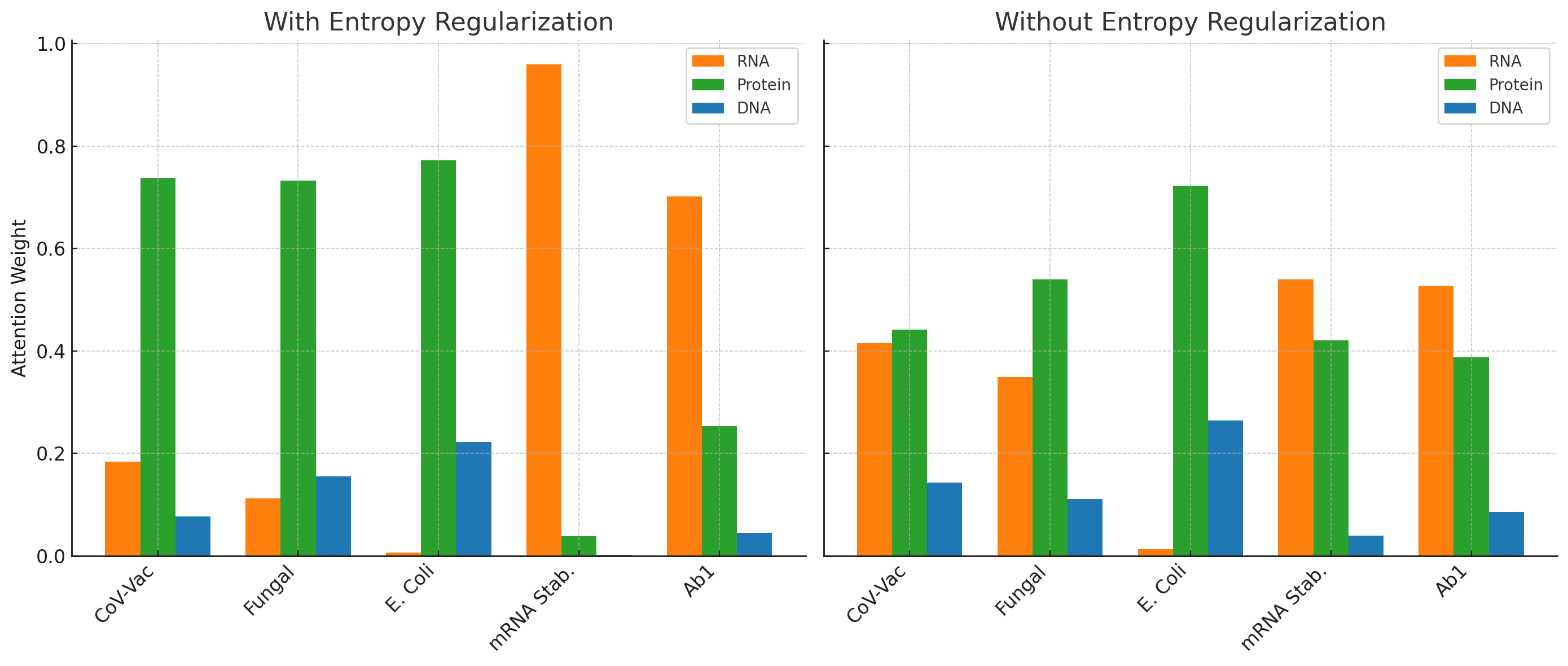}
        \vspace{-2em}
        \caption{Modality-wise attention weights with and without entropy regularization from MIL based fusion model}
        \label{fig:attention_entropy}
\end{figure}

\vspace{-0.3em}
\paragraph{Dataset}
We evaluate \textsc{BioLangFusion} on five biologically diverse datasets covering both regression and classification tasks. Each dataset consists of mRNA sequences paired with phenotype-relevant labels, spanning applications such as vaccine design, expression profiling, stability analysis, and antibody prediction. These datasets differ in sequence length, sample size, and prediction targets. Full statistics and details are provided in Appendix~\ref{sec:appen_datastat}.
\vspace{-0.5em}
\paragraph{Experimental Setup}
We begin by extracting DNA, RNA, and protein embeddings using pretrained FMs— \emph{Nucleotide Transformer}, \emph{RNA-FM}, and \emph{ESM-2 (8M)} (see Table~\ref{tab:foundation_models}). These embeddings are then fused using our proposed strategies to produce a unified multimodal representation, which is passed to a downstream prediction head.  All fusion methods share the same fixed TextCNN prediction head to ensure a fair comparison (see Appendix~\ref{sec:experiments_architecture} for details).
\vspace{-0.6em}
\paragraph{Experimental Results}
\begin{table}[h]
  \centering
  \scriptsize
  \resizebox{\columnwidth}{!}{%
    \begin{tabular}{@{}lccccc@{}}
      \toprule
      Encoding               & CoV-Vac & Fungal  & E.\,Coli & mRNA Stab. & Ab1 \\
      \toprule
      Evo \citep{lin2023evolutionary}    & 0.653 & 0.579 & 42.556 & 0.403  & 0.360 \\
      SpliceBERT \cite{chen2024self}&0.802&0.778&48.455&0.522&0.718 \\
      ESM-2 (650M)   & 0.825 & 0.734 & 46.348 & 0.536 & 0.679\\
      ESM-2 (3B)    & 0.772 & 0.721 & 46.208 & 0.537 & 0.700\\
      \midrule
      \emph{ESM-2 (8M)}& 0.806  & 0.695  & 49.017   & 0.539     & 0.711 \\
      \emph{RNA-FM }      & 0.841  & 0.767  & 52.949   & 0.553     & 0.743 \\
      \emph{Nucleotide Transformer }               & 0.780  & 0.804& 41.292 & 0.530     & 0.732 \\
      \midrule
      \textbf{Concatenation}         & 0.831  & 0.805  & 50.280   & 0.539     & 0.764 \\
      \textbf{MIL + Entropy}         & \textbf{0.864} & \textbf{0.824} & 52.107 & \textbf{0.563} & 0.760 \\
     \textbf{ Cross Attention}       & 0.828  & 0.812  & \textbf{53.932} & 0.550 & \textbf{0.765} \\
      \bottomrule
    \end{tabular}%
  }
 \caption{Performance of  multimodal fusion variants of \textsc{BioLangFusion} vs. single-modal baselines. Spearman correlation is used for regression; accuracy for classification (E. coli). We use  \emph{Nucleotide Transformer}, \emph{RNA-FM}, and \emph{ESM-2 (8M)} for fusion}
\label{tab:main_results}
\end{table}
\vspace{-0.5em}
We report model performance in Table~\ref{tab:main_results}.
Across all the tasks, fusion-based models consistently outperform the best single-modality baselines. While naive concatenation already benefits from multi-modal information, attention pooling further improves performance by learning task-specific modality weights. This allows the model to prioritize the most informative modality for each input, rather than treating all modalities equally. Interestingly, the cross-attention fusion method achieves the best performance on the \textit{E. coli Proteins} dataset, but generally underperforms compared to MIL-based attention pooling on most other tasks. This discrepancy suggests two possibilities: either token-level cross-modal interactions are particularly beneficial for certain tasks such as protein abundance classification, or the current datasets may be too limited in size to fully exploit the representational capacity of cross-attention mechanisms. 
\vspace{-0.5em}
\paragraph{Explaining Modality Contributions via Attention}\label{sec:interpretability}
To better understand how BioLangFusion integrates complementary signals from DNA, mRNA, and protein modalities, we visualize the learned attention weights assigned to each modality across datasets. These attention scores reflect the importance attributed to each modality in generating fused representations for downstream predictions. Figure~\ref{fig:attention_entropy} shows modality-wise attention scores from the entropy-regularized model. Clear biological trends emerge—for instance, in the \textit{mRNA Stab.} task, the model emphasizes mRNA embeddings, consistent with the role of post-transcriptional features in degradation. In contrast, predictions of \textit{E. coli} protein abundance exhibit stronger attention toward protein embeddings. We then compare these scores with attention score obtained from training without entropy. In the absence of entropy regularization, attention weights tend to be diffuse and less decisive.   In contrast, entropy‐regularized attention yields sharp, task‐specific modality weights, enhancing interpretability and revealing each pretrained FM’s contribution to the downstream decisions (See Appendix~\ref{sec:ablation}).
\vspace{-1em}

\vspace{-1em}
\section{Conclusion}
\vspace{-0.5em}
In this paper, we explore strategies to fuse pretrained DNA, mRNA, and protein foundation models, capturing their biological interconnectivity. 
We introduce \textsc{BioLangFusion}, a lightweight, plug-and-play suite of fusion strategies for integrating pretrained DNA, mRNA, and protein FMs. 
It aligns embeddings at the codon level and incorporates biologically informed fusion mechanisms—including codon-aware concatenation, entropy-regularized attention pooling, and cross-modal multi-head attention. Across five molecular property prediction tasks, BioLangFusion consistently outperforms unimodal baselines, demonstrating the synergistic power of pretrained unimodal FMs when fused effectively. Moreover, our approach provides interpretability through modality attention weights, revealing the biological relevance of each modality across tasks. BioLangFusion requires no end-to-end retraining or architectural changes, offering a scalable and accessible framework for multi-modal modeling in biology.

\clearpage
\bibliography{reference}
\bibliographystyle{icml2025}
\newpage

\newpage
\appendix
\onecolumn
\section{Appendix}
\subsection{Related Work}
\label{sec:related}


\paragraph{Unimodal biomolecular language models.}
Recent advances in biological language modeling have produced powerful unimodal models trained on large-scale corpora of DNA, RNA, and protein sequences. In genomics, transformer-based architectures such as DNABERT~\citep{ji2021dnabert}, DNABERT-2~\citep{liu2023dnabert2}, and Nucleotide Transformer~\citep{dalla2024nucleotide} have demonstrated strong performance on tasks like regulatory element prediction and variant prioritization, while complementary models such as HyenaDNA~\citep{nguyen2023hyenadna} and Evo~\citep{nguyen2024evo} use long-range convolution or structured state space dynamics to enhance scalability and contextual understanding. For protein sequences, models including ProtTrans~\citep{elnaggar2021prottrans}, ESM family of models~\citep{lin2022esm}, Ankh~\citep{elnaggar2023ankh}, and ProGen2~\citep{madani2023progen2} have achieved state-of-the-art results in structure prediction, function classification, and mutational fitness estimation via zero-shot transfer. In RNA, models such as RNA-FM~\citep{chen2022rnafm}, SpliceBERT~\citep{wang2023splicebert}, and UTR-LM~\citep{li2023utrlm} focus on various non-coding regions, while recent mRNA-specific models like CodonBERT~\citep{zhang2023codonbert} and HELM~\citep{yazdani2024helm} have also been proposed.

\paragraph{Multi-modal integration of bio-LMs.}
While multimodal learning has gained traction in many domains, integrating biological modalities—DNA, RNA, and proteins—remains an underexplored area. Most existing approaches rely on rudimentary fusion strategies such as early feature concatenation~\citep{multiomics_concat} or late-stage prediction averaging~\citep{ceviche2024}, which fail to harness the expressive power of contextual embeddings from pretrained biological language models. Moreover, these methods overlook inter-modality alignment and do not scale well to more than two modalities due to dimensional constraints imposed by naive concatenation.
In the protein domain, Tranception~\citep{tranception2023} combines language model outputs with multiple sequence alignment (MSA) features via blockwise attention, but it is limited to protein-only modeling and ignores transcriptional or genomic context. Recent work~\citep{evo-transfer} has explored direct transfer of pretrained DNA and protein models to mRNA-specific tasks, demonstrating promising cross-modal generalization, though without modeling all modalities jointly. Life-Code~\citep{lifecode_paper}, in contrast, proposes a more holistic multi-omics architecture inspired by the central dogma, integrating modalities through reverse translation and codon-aware modeling, but requires pre-training bespoke foundation models from scratch.

To date, few approaches aim to unify DNA, RNA, and protein representations in a compute-efficient framework. Additionally, biologically meaningful alignment strategies—such as codon-level mappings and transcript-aware position encoding—remain largely absent, limiting the capacity to model cross-modal dependencies embedded in molecular biology’s central dogma.

\subsection{Dataset}\label{sec:appen_datastat}
We evaluate our method on five molecular property prediction datasets. The Ab1 dataset is sourced from \citet{yazdani2024helm}, while the remaining four are obtained from \citet{li2023codonbert}.

\begin{itemize}
    \item \textbf{CoV-Vac} \citep{leppek2022combinatorial}: SARS-CoV-2 vaccine degradation dataset comprising mRNA sequences engineered for optimized structural features, stability, and translational efficiency in vaccine development contexts.
    
    \item \textbf{Fungal} \citep{wint2022kingdom}: Expression dataset containing protein-coding and tRNA genes extracted from a wide range of fungal genomes, annotated with corresponding expression levels.
    
    \item \textbf{E. coli} \citep{ding2022mpepe}: Experimental dataset of protein expression in \textit{E. coli}, with expression levels categorized into low, medium, and high classes.
    
    \item \textbf{mRNA Stab.} \citep{diez2022icodon}: A large dataset containing mRNA sequences from human, mouse, frog, and fish, annotated with experimentally measured transcript stability scores.
    
    \item \textbf{Ab1} \citep{yazdani2024helm}: A collection of antibody-encoding mRNA sequences labeled with quantitative protein expression levels.
\end{itemize}

We follow the train/test splits provided by \citet{li2023codonbert}. To ensure compatibility with pretrained language models—most of which support input lengths up to ~1000 tokens—we truncate RNA sequences exceeding this length. Dataset statistics are summarized in Table~\ref{tab:data-stats}.

\begin{table}[h]
\centering
\begin{tabular}{llllll}
\toprule
\textbf{Dataset} & \textbf{Max Length} & \#mRNA (raw) & \#mRNA (used) & \textbf{Target} & \textbf{Task} \\
\midrule
CoV-Vac        & 81    & 2400  & 2400  & Degradation & Regression \\
Fungal         & 3063  & 7056  & 3138  & Expression  & Regression \\
E. coli        & 3000  & 6348  & 4450  & Expression  & Classification \\
mRNA Stab. & 3066  & 41123 & 23929 & Stability    & Regression \\
Ab1            & 1203  & 723   & 723   & Expression  & Regression \\
\bottomrule
\end{tabular}
\caption{Summary of the datasets used in our experiments, including the maximum input sequence length, number of available mRNA sequences before and after preprocessing, the type of molecular property being predicted, and the corresponding task type (regression or classification).}

\label{tab:data-stats}
\end{table}

The foundational models used for each modality and their embedding dimensionalities are shown in Table~\ref{tab:foundation_models}.

\begin{table}[H]
  \centering
  \small
  \begin{tabular}{@{}lllr@{}}
    \toprule
    \textbf{Modality} & \textbf{Model} & \textbf{Version} & \textbf{Embedding Dim.} \\
    \midrule
    RNA     & RNA-FM             & \texttt{rna\_fm\_t12}                   & 640 \\
    DNA     & Nucleotide Transformer & \texttt{nucleotide-transformer-v2-100m-multi-species} & 4,107 \\
    Protein & ESM-2              & \texttt{esm2\_t6\_8M\_UR50D}           & 320 \\
    \bottomrule
  \end{tabular}
  \caption{Pretrained foundation models used to encode  each modality, listing the exact version and embedding dimension used as input to the fusion framework.}
  \label{tab:foundation_models}
\end{table}

\subsection{Ablation study}\label{sec:ablation}
To better understand the contribution of individual components and architectural choices in BioLangFusion, we conduct a comprehensive ablation study. We begin by examining the effect of the entropy regularization term on the MIL-based attention fusion (Section ~\ref{sec:mil}). This regularization encourages non-uniform attention distributions and, as shown in Section~\ref{sec:interpretability}, improves both performance and interpretability in low-data regimes.

Next, we assess the impact of using shared versus modality-specific projection layers in the MIL attention mechanism. In our default setup, each modality uses its own projection before attention calculation; we ablate this design by sharing a single projection across all modalities (in Eq.~\ref{eq:gated_attention}) to test whether distinct projections are necessary for capturing modality-specific characteristics.

We further explore token-level attention by removing the mean pooling step  prior to computing attention scores. This variant produces per-token attention vectors for each modality while MIL base attention produce one attention score per sequence, allowing the model to learn finer-grained importance distributions along the sequence.

Finally, to assess whether simple concatenation works without modality alignment (\ref{sec:modality_align}), we experiment with a naive strategy that first projects all embeddings to a common feature dimension and concatenates them along the sequence length axis.

\begin{table}[H]
  \centering
    \begin{tabular}{@{}lccccc@{}}
      \toprule
      Method & CoV-Vac & Fungal & E.\,Coli & mRNA Stab. & Ab1 \\
      \midrule
      MIL without entropy          & 0.854 & \textbf{0.826} & 51.960 & 0.556 & 0.753 \\
      MIL with shared projection (with entropy)   & 0.859 & 0.813 & 50.140 & \textbf{0.563} & 0.755 \\
      Token level attention        & 0.835 & 0.764 & 44.100 & 0.553 & \textbf{0.770} \\
      Vanilla concatenation (without codon level alignment) &0.818& 0.807&46.208&0.537&0.754\\
      \midrule
      MIL + entropy                & \textbf{0.864} & 0.824 & \textbf{52.107} & \textbf{0.563} & 0.760 \\
      \bottomrule
    \end{tabular}
  \caption{Performance comparison of attention variants across selected datasets (rounded to 3 decimal places).}
  \label{Tab:ablation}
\end{table}

As shown in Table ~\ref{Tab:ablation}, entropy regularizer improves prediction performance. Moverove, having modality specific projection layer when we calculate the gated attention score improves over having shared projection layers accross modality. Finally, token level attention brings performance gain only on one (Ab1) out of the five data set.

\subsection{Network architecture and optimization parameters}\label{sec:experiments_architecture}
In Figure \ref{fig:method}, we present the model architecture for the three fusion methods::
\begin{figure}[H]
    \centering
    \includegraphics[scale = 0.8]{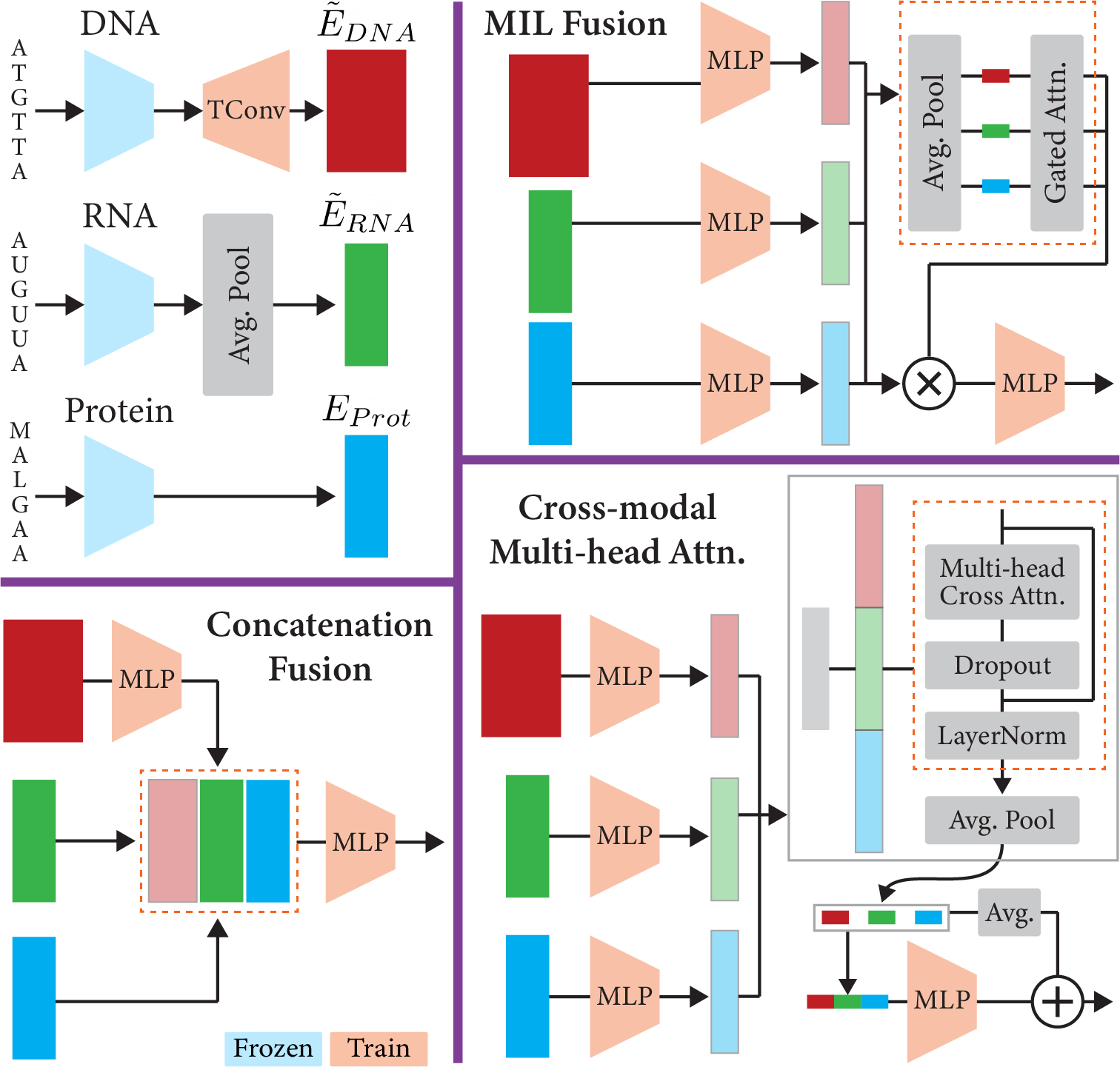}
    \caption{
    Model architectures for the three BioLangFusion fusion strategies.
Codon-aware concatenation: DNA embeddings are upsampled via transposed convolution and mRNA embeddings downsampled via non-overlapping mean pooling to align with the protein framing; the three aligned embeddings are then concatenated at each codon position and projected before entering the TextCNN head.
 Entropy-regularized gated attention (MIL): After alignment, each modality is projected to a shared latent space; a gated attention mechanism with an entropy regularizer computes modality weights per sequence, producing a weighted sum fused embedding that feeds into the TextCNN head.
 Cross-modal multi-head attention: All modality embeddings are projected, concatenated along the temporal axis to form a joint context, and each modality attends to this context via multi-head attention; the updated embeddings are merged with residual averaging and layer normalization before the TextCNN head.
}
    \label{fig:method}
\end{figure}

We train all models using the Adam optimizer with a learning rate of $3 \times 10^{-5}$ and weight decay of $1 \times 10^{-5}$. Early stopping is employed with a patience of 20 epochs, and the learning rate is scheduled using \texttt{ReduceLROnPlateau} with a patience of 5. For regression tasks, we use mean squared error (MSE) loss; for classification, we use cross-entropy. The entropy regularization weight $\lambda$ is selected from the set $\{0.01, 0.5, 1\}$ via validation performance. All experiments are conducted with a batch size of 32 on a single GPU. The detailed network architecture and training hyperparameters are provided in List~\ref{alg:biofusion_architecture} and List ~\ref{alg:biofusion_training}.

\begin{algorithm*}
\dirtree{%
.0 .
.1 Input modality embeddings.
.2 Protein embedding \dotfill \{ESM-2 (320-dim)\}.
.2 RNA embedding \dotfill \{RNA-FM T12 (640-dim)\}.
.2 DNA embedding \dotfill \{Nucleotide Transformer v2 (4107-dim)\}.
.1 Temporal alignment.
.2 RNA downsampling \dotfill \{AvgPool1D (kernel=3, stride=3)\}.
.2 DNA upsampling \dotfill \{ConvTranspose1D (kernel=3, stride=2, padding=2)\}.
.2 Padding \dotfill \{sequence length aligned to protein\}.
.1 Projection layers.
.2 Per-modality projection \dotfill \{Linear $\rightarrow$ 600-dim\}.
.2 Activation function \dotfill \{tanh\}.
.1 Attention fusion.
.2 Attention dimension \dotfill \{100\}.
.2 Gating dimension \dotfill \{100\}.
.2 Softmax temperature $\tau$ \dotfill \{learned, clamped to [0.02, 20.0]\}.
.2 Fusion operation \dotfill \{weighted sum of modality embeddings\}.
.1 TextCNN prediction head.
.2 Conv1D kernel sizes \dotfill \{3, 4, 5\}.
.2 Conv1D output channels \dotfill \{1280\}.
.2 Activation function \dotfill \{ReLU\}.
.2 Global max pooling.
.2 Dropout \dotfill \{0.2\}.
.2 Fully connected layer \dotfill \{output task-specific prediction\}.
}
\caption{BioLangFusion architecture:MIL based attention fusion and TextCNN head. \label{alg:biofusion_architecture}}
\end{algorithm*}

\begin{algorithm*}
\dirtree{%
.0 .
.1 Input modality embeddings.
.2 Protein embedding \dotfill \{ESM-2 (320-dim)\}.
.2 RNA embedding \dotfill \{RNA-FM T12 (640-dim)\}.
.2 DNA embedding \dotfill \{Nucleotide Transformer v2 (4107-dim)\}.
.1 Temporal alignment.
.2 RNA downsampling \dotfill \{AvgPool1D (kernel=3, stride=3)\}.
.2 DNA upsampling \dotfill \{ConvTranspose1D (kernel=3, stride=2, padding=2)\}.
.2 Padding \dotfill \{sequence length aligned to protein\}.
.1 Projection layers.
.2 Per-modality projection \dotfill \{Linear\(\to600\)-dim, tanh activation\}.
.1 Cross-modal attention.
.2 Build joint context \dotfill \{concat along time \(\to3T'\times600\)\}.
.2 MultiHeadAttention \dotfill \{4 heads, dropout 0.1\}.
.2 Residual + LayerNorm \dotfill \{per modality\}.
.1 Fusion.
.2 Concatenate streams \dotfill \{shape \(T'\times1800\)\}.
.2 Linear fusion \dotfill \{\(1800\to600\)-dim\}.
.2 Residual average + LayerNorm.
.1 TextCNN prediction head.
.2 Conv1D kernels \dotfill \{3, 4, 5\}, 100 channels.
.2 Activation \dotfill \{ReLU\}.
.2 Global max pooling.
.2 Dropout \dotfill \{0.2\}.
.2 Fully connected layer \dotfill \{task-specific output\}.
}
\caption{BioLangFusion architecture: cross-modal multi-head attention fusion and TextCNN head.}
\label{alg:cross_attention_architecture}
\end{algorithm*}

\begin{algorithm*}
\dirtree{%
.0 .
.1 Objective function.
.2 Regression task \dotfill \{MSE + $\lambda$Entropy\}.
.2 Classification task \dotfill \{CrossEntropy + $\lambda$Entropy\}.
.1 Optimization algorithm \dotfill \{Adam\}.
.1 Learning rate \dotfill \{$3 \times 10^{-5}$\}.
.1 Weight decay \dotfill \{$1 \times 10^{-5}$\}.
.1 Learning rate scheduler \dotfill \{ReduceLROnPlateau (patience=5)\}.
.1 Early stopping \dotfill \{patience = 20 epochs\}.
.1 Batch size \dotfill \{32\}.
.1 Max training epochs \dotfill \{500\}.
.1 Evaluation metrics.
.2 Regression \dotfill \{Spearman\}.
.2 Classification \dotfill \{Accuracy\}.
}
\caption{Training setup and optimization details}
\label{alg:biofusion_training}
\end{algorithm*}





\end{document}